\documentclass[10pt,twocolumn,letterpaper]{article}

\usepackage{cvpr}
\usepackage{times}
\usepackage{epsfig}
\usepackage{graphicx}
\usepackage{amsmath}
\usepackage{amssymb}
\usepackage{tabu}

\usepackage{pgfplotstable}
\usepackage{pgfplots}
\usepackage{graphicx,dblfloatfix}
\usepackage{epstopdf}
\usepackage{subcaption}
\usepackage{longtable,ltxtable,booktabs}
\usepackage{enumitem}
\usepackage{pgfplotstable}
\usepackage{pgfplots}
\usepackage[noend]{algpseudocode}
\usepackage[linesnumbered,ruled]{algorithm2e}

\usepackage[pagebackref=true,breaklinks=true,letterpaper=true,colorlinks,bookmarks=false]{hyperref}

 \cvprfinalcopy 

\def\cvprPaperID{692} 

\ifcvprfinal\fi

\begin{document}
\title{Joint Geometrical and Statistical Alignment for Visual Domain Adaptation}

\author{Jing Zhang, Wanqing Li, Philip Ogunbona\\
Advanced Multimedia Research Lab, University of Wollongong, Australia\\
{\tt\small jz960@uowmail.edu.au,}
{\tt\small wanqing@uow.edu.au,}
{\tt\small philipo@uow.edu.au}\\
}



\maketitle
\begin{abstract}
This paper presents a novel unsupervised domain adaptation method for cross-domain visual recognition. We propose a unified framework that reduces the shift between domains both statistically and geometrically, referred to as Joint Geometrical and Statistical Alignment (JGSA). Specifically, we learn two coupled projections that project the source domain and target domain data into low-dimensional subspaces where the geometrical shift and distribution shift are reduced simultaneously. The objective function can be solved efficiently in a closed form. Extensive experiments have verified that the proposed method significantly outperforms several state-of-the-art domain adaptation methods on a synthetic dataset and three different real world cross-domain visual recognition tasks.

\end{abstract}

\section{Introduction}
A basic assumption of statistical learning theory is that the training and test data are drawn from the same distribution. Unfortunately, this assumption does not hold in many applications. For example, in visual recognition, the distributions between training and test can be discrepant due to the environment, sensor type, resolution, and view angle. In video based visual recognition, more factors are involved in addition to those in image based visual recognition. For example, in action recognition, the subject, performing style, and performing speed increase the domain shift further. Labelling data is labour intensive and expensive, thus it is impractical to relabel a large amount of data in a new domain. Hence, a realistic strategy, domain adaptation, can be used to employ previous labeled source domain data to boost the task in the new target domain. Based on the availability of target labeled data, domain adaptation can be generally divided into semi-supervised and unsupervised domain adaptation. The semi-supervised approach requires a certain amount of labelled training samples in the target domain and the unsupervised one requires none labelled data. 
However, in both semi-supervised and unsupervised domain adaptation, sufficient unlabeled target domain data are required. In this paper, we focus on unsupervised domain adaptation which is considered to be more practical and challenging.

The most commonly used domain adaptation approaches include instance-based adaptation, feature representation adaptation, and classifier-based adaptation~\cite{Pan2010,Shao2015}.
In unsupervised domain adaptation, as there is no labeled data in the target domain, the classifier-based adaptation is not feasible. Alternatively, we can deal with this problem by minimizing distribution divergence between domains as well as the empirical source error~\cite{Ben-David2010}.
It is generally assumed that the distribution divergence can be compensated either by an instance based adaptation method, such as reweighting samples in the source domain to better match the target domain distribution, or by a feature transformation based method that projects features of two domains into another subspace with small distribution shift. 
The instance-based approach requires the strict assumptions~\cite{Pan2010,Margolis2011} that 1) the conditional distributions of source and target domain are identical, and 2) certain portion of the data in the source domain can be reused for learning in the target domain through reweighting.
While the feature transformation based approach relaxes these assumptions, and only assumes that there exists a common space where the distributions of two domains are similar. This paper follows the feature transformation based approach.

Two main categories of feature transformation methods are identified~\cite{Yang2015} among the literature, namely data centric methods and subspace centric methods. The data centric methods seek a unified transformation that projects data from two domains into a domain invariant space to reduce the distributional divergence between domains while preserving data properties in original spaces, such as~\cite{Pan2011,Long2013,Long2014,Ghifary2016}. The data centric methods only exploit shared feature in two domains, which will fail when the two different domains have large discrepancy, because there may not exist such a common space where the distributions of two domains are the same and the data properties are also maximumly preserved in the mean time. For the subspace centric methods, the domain shift is reduced by manipulating the subspaces of the two domains such that the subspace of each individual domain all contributes to the final mapping~\cite{Gong2012,Fernando2013,Fernando2015}. Hence, the domain specific features are exploited. For example, Gong et al.~\cite{Gong2012} regard two subspaces as two points on Grassmann manifold, and find points on a geodesic path between them as a bridge between source and target subspaces. Fernando et al. ~\cite{Fernando2013} align source and target subspaces directly using a linear transformation matrix. However, the subspace centric methods only manipulate on the subspaces of the two domains without explicitly considering the distribution shift between projected data of two domains. The limitations of both data centric and subspace centric methods will be illustrated on a synthetic dataset in Section~\ref{sec:synthetic}.

In this paper, we propose a unified framework that reduces the distributional and geometrical divergence between domains simultaneously by exploiting both the shared and domain specific features. Specifically, we learn two coupled projections to map the source and target data into respective subspaces.
After the projections, 1) the variance of target domain data is maximized to preserve the target domain data properties, 2) the discriminative information of source data is preserved to effectively transfer the class information, 3) both the marginal and conditional distribution divergences between source and target domains are minimized to reduce the domain shift statistically, and 4) the divergence of two projections is constrained to be small to reduce domain shift geometrically. 

Hence, different from data centric based methods, we do not require the strong assumption that a unified transformation can reduce the distribution shift while preserving
the data properties. Different from subspace centric based methods, we not only reduce the shift of subspace geometries but also reduce the distribution shifts of two domains.
In addition, our method can be easily extended to a kernelized version to deal with the situations where the shift between domains are nonlinear. The objective function can be solved efficiently in a closed form. The proposed method has been verified through comprehensive experiments on a synthetic dataset and three different real world cross-domain visual recognition tasks: object recognition (Office, Caltech-256), hand-written digit recognition (USPS, MNIST), and RGB-D-based action recognition (MSRAction3DExt, G3D, UTD-MHAD, and MAD).
%
%

\section{Related Work}
\subsection{Data centric approach}
Pan et al.~\cite{Pan2011} propose the transfer component analysis (TCA) to learn some transfer components across domains in RKHS using Maximum Mean Discrepancy (MMD)~\cite{Gretton2012}. TCA is a typical data centric approach that finds a unified transformation $\phi(\cdot)$ that projects data from two domains into a new space to reduce the discrepancy. In TCA, the authors aim to minimize the distance between the sample means of the source and target data in the k-dimensional embeddings while preserving data properties in original spaces. Joint distribution analysis (JDA)~\cite{Long2013} improves TCA by considering not only the marginal distribution shift but also the conditional distribution shift using the pseudo labels of target domain. Transfer joint matching (TJM)~\cite{Long2014} improves TCA by jointly reweighting the instances and finding the common subspace. Scatter component analysis (SCA)~\cite{Ghifary2016} takes the between and within class scatter of source domain into consideration. However, these methods require a strong assumption that there exist a unified transformation to map source and target domains into a shared subspace with small distribution shift. 
\subsection{Subspace Centric Approach}
As mentioned, subspace centric approach can address the issue of data centric methods that only exploit common features of two domains. Fernando et al.~\cite{Fernando2013} proposed a subspace centric method, namely Subspace Alignment (SA). The key idea of SA is to align the source basis vectors ($A$) with the target one ($B$) using a transformation matrix $M$. $A$ and $B$ are obtained by PCA on source and target domains, respectively. 
Hence, they do not assume that there exist a unified transformation to reduce the domain shifts.
However, the variance of projected source domain data will be different from that of target domain after mapping the source subspace using a linear map because of the domain shift. In this case, SA fails to minimize the distributions between domains after aligning the subspaces. In addition, SA cannot deal with situations where the shift between two subspaces are nonlinear.
Subspace distribution alignment (SDA)~\cite{Sun2015} improves SA by considering the variance of the orthogonal principal components. However, the variances are considered based on the aligned subspaces. Hence, only the magnitude of each eigen direction is changed which may still fail when the domain shift is large. This has been validated by the illustration of synthetic data in Figure~\ref{fig:toy} and the experiment results on real world datasets. 

\section{Joint Geometrical and Statistical Alignment}
This section presents the Joint Geometrical and Statistical Alignment (JGSA) method in detail.
\subsection{Problem Definition}
We begin with the definitions of terminologies.
%
The source domain data denoted as $X_s\in{\mathbb{R}^{D\times n_s}}$ are draw from distribution $P_s(X_s)$ and the target domain data denoted as $X_t\in{\mathbb{R}^{D\times n_t}}$ are draw from distribution $P_t(X_t)$, where D is the dimension of the data instance, $n_s$ and $n_t$ are number of samples in source and target domain respectively. 
We focus on the unsupervised domain adaptation problem. 
In unsupervised domain adaptation, there are sufficient labeled source domain data, $\mathcal{D}_s = \{(\mathbf{x}_i,y_i)\}_{i=1}^{n_s}$, $\mathbf{x}_i\in{\mathbb{R}^D}$, and unlabeled target domain data, $\mathcal{D}_t = \{(\mathbf{x}_j)\}_{j=1}^{n_t}$, $\mathbf{x}_j\in{\mathbb{R}^D}$, in the training stage. We assume the feature spaces and label spaces between domains are the same: $\mathcal{X}_s = \mathcal{X}_t$ and $\mathcal{Y}_s = \mathcal{Y}_t$.
Due to the dataset shift, $P_s(X_s)\neq{P_t(X_t)}$. 
Different from previous domain adaptation methods, we do not assume that there exists a unified transformation $\phi(\cdot)$ such that $P_s(\phi(X_s))=P_t(\phi(X_t))$ and $P_s(Y_s|\phi(X_s))={P_t(Y_t|\phi(X_s))}$, since this assumption becomes invalid when the dataset shift is large. 

\subsection{Formulation}
To address limitations of both data centric and subspace centric methods, 
the proposed framework (JGSA) reduces the domain divergence both statistically and geometrically by exploiting both shared and domain specific features of two domains. 
The JGSA is formulated by finding two coupled projections (A for source domain, and B for target domain) to obtain new representations of respective domains, such that 1) the variance of target domain is maximized, 2) the discriminative information of source domain is preserved, 3) the divergence of source and target distributions is small, and 4) the divergence between source and target subspaces is small.

\vspace{-0.5em}
\subsubsection{Target Variance Maximization}
To avoid projecting features into irrelevant dimensions, we encourage the variances of target domain is maximized in the respective subspaces. Hence, the variance maximization can be achieved as follows
\begin{equation}
\max_{B} Tr(B^TS_tB)
\label{eqt:target}
\end{equation}
where 
\begin{equation}
S_t=X_tH_tX_t^T
\label{eqt:St}
\end{equation}
is the target domain scatter matrix, $H_t = I_t-\frac{1}{n_t}1_t1_t^T$ is the centering matrix, $1_t\in{\mathbb{R}^{n_t}}$ is the column vector with all ones.

\vspace{-0.5em}
\subsubsection{Source Discriminative Information Preservation} 
Since the labels in the source domain are available, we can employ the label information to constrain the new representation of source domain data to be discriminative.
\begin{equation}
\max_{A} Tr(A^TS_bA)
\label{eqt:between}
\end{equation}
\begin{equation}
\min_{A} Tr(A^TS_wA)
\label{eqt:within}
\end{equation}
where $S_w$ is the within class scatter matrix, and $S_b$ is the between class scatter matrix of the source domain data, which are defined as follows,
\begin{equation}
S_w = \sum_{c=1}^C X_s^{(c)}H_s^{(c)}(X_s^{(c)})^T
\end{equation}
\begin{equation}
S_b = \sum_{c=1}^C n_s^{(c)}(m_s^{(c)}-\bar{m}_s)(m_s^{(c)}-\bar{m}_s)^T
\end{equation}
where $X_s^{(c)}\in {\mathbb{R}^{D\times n_s^{(c)}}}$ is the set of source samples belonging to class $c$, $m_s^{(c)} = \frac{1}{n_s^{(c)}}\sum_{i=1}^{n_s^{(c)}}x_i^{(c)}$, $\bar{m}_s = \frac{1}{n_s}\sum_{i=1}^{n_s}x_i$, $H_s^{(c)} = I_s^{(c)}-\frac{1}{n_s^{(c)}}1_s^{(c)}(1_s^{(c)})^T$ is the centering matrix of data within class $c$, $I_s^{(c)}\in{\mathbb{R}^{n_s^{(c)}\times n_s^{(c)}}}$ is the identity matrix, $1_s\in{\mathbb{R}^{n_s^{(c)}}}$ is the column vector with all ones, $n_s^{(c)}$ is the number of source samples in class $c$.

\vspace{-0.5em}
\subsubsection{Distribution Divergence Minimization}
We employ the MMD criteria~\cite{Gretton2012,Pan2011,Long2013} to compare the distributions between domains, which computes the distance between the sample means of the source and target data in the k-dimensional embeddings,
\begin{equation}
\min_{A,B} \|\frac{1}{n_s}\sum_{\mathbf{x}_i\in{X_s}}A^T\mathbf{x}_i-\frac{1}{n_t}\sum_{\mathbf{x}_j\in{X_t}}B^T\mathbf{x}_j\|^2_F 
\label{eqt:MMDterm1}
\end{equation}
Long et al.~\cite{Long2013} has been proposed to utilize target pseudo labels predicted by source domain classifiers for representing the class-conditional data distributions in the target domain. Then the pseudo labels of target domain are iteratively refined to reduce the difference in conditional distributions between two domains further. We follow their idea to minimize the conditional distribution shift between domains,
\begin{align}
\min_{A,B} \sum_{c=1}^C\|\frac{1}{n^{(c)}_s}\sum_{\mathbf{x}_i\in{X_s^{(c)}}}A^T\mathbf{x}_i-\frac{1}{n^{(c)}_t}\sum_{\mathbf{x}_j\in{X_t^{(c)}}}B^T\mathbf{x}_j\|^2_F
\label{eqt:MMDterm2}
\end{align}
Hence, by combining the marginal and conditional distribution shift minimization terms, the final distribution divergence minimization term can be rewritten as 
\begin{align}
\min_{A,B} Tr\bigg([\begin{matrix}
 A^T & B^T
 \end{matrix}] 
\Bigg[ \begin{matrix}
  M_s & M_{st} \\
  M_{ts}& M_t
 \end{matrix}\Bigg] \Bigg[ \begin{matrix}
  A \\
  B
 \end{matrix} \Bigg] \big)
 \label{eqt:MMD}
\end{align}
where 
\begin{small}
\begin{align}
\begin{split}
M_s = X_s(L_s+\sum_{c=1}^C L_s^{(c)})X_s^T, \quad L_s = \frac{1}{n_s^2}1_s1_s^T,\\ (L_s^{(c)})_{ij}= \begin{cases}
\frac{1}{(n^{(c)}_s)^2} & \quad  \mathbf{x}_i,\mathbf{x}_j\in{X_s^{(c)}}\\
0      & \quad \text{otherwise } 
\end{cases}
\end{split}
\label{eqt:Ms}
\end{align}
\vspace{-1em}
\begin{align}
\begin{split}
M_t = X_t(L_t+\sum_{c=1}^C L_t^{(c)})X_t^T, \quad L_t = \frac{1}{n_t^2}1_t1_t^T,\\ (L_t^{(c)})_{ij}= \begin{cases}
\frac{1}{(n^{(c)}_t)^2} & \quad  \mathbf{x}_i,\mathbf{x}_j\in{X_t^{(c)}}\\
0      & \quad \text{otherwise } 
\end{cases}
\end{split}
\label{eqt:Mt}
\end{align}
\vspace{-1em}
\begin{align}
\begin{split}
M_{st} = X_s(L_{st}+\sum_{c=1}^C L_{st}^{(c)})X_t^T, \quad L_{st} = -\frac{1}{n_sn_t}1_s1_t^T, \\ 
(L_{st}^{(c)})_{ij}= \begin{cases}
-\frac{1}{n^{(c)}_s n^{(c)}_t} & \quad   
    	\mathbf{x}_i\in{X_s^{(c)}},\mathbf{x}_j\in{X_t^{(c)}} \\
0      & \quad \text{otherwise } 
\end{cases}
\end{split}
\label{eqt:Mst}
\end{align}
\vspace{-1em}
\begin{align}
\begin{split}
M_{ts} = X_t(L_{ts}+\sum_{c=1}^C L_{ts}^{(c)})X_s^T, \quad L_{ts} = -\frac{1}{n_sn_t}1_t1_s^T, \\ 
(L_{ts}^{(c)})_{ij}= \begin{cases}
-\frac{1}{n^{(c)}_s n^{(c)}_t} & \quad  
    	\mathbf{x}_j\in{X_s^{(c)}},\mathbf{x}_i\in{X_t^{(c)}} \\
0      & \quad \text{otherwise } 
\end{cases}
\end{split}
\label{eqt:Mts}
\end{align}
\end{small}
Note that this is different from TCA and JDA, because we do not use a unified subspace because there may not exist such a common subspace where the distributions of two domains are also similar.

\subsubsection{Subspace Divergence Minimization}
Similar to SA~\cite{Fernando2013}, we also reduce the discrepancy between domains by moving closer the source and target subspaces. As mentioned, an additional transformation matrix $M$ is required to map the source subspace to the target subspace in SA. However, we do not learn an additional matrix to map the two subspaces. Rather, we optimize A and B simultaneously, such that the source class information and the target variance can be preserved, and the two subspaces move closer in the mean time.
We use following term to move the two subspaces close:
\begin{equation}
\min_{A,B}\|A-B\|^2_F
\label{eqt:SA}
\end{equation} 
By using term~(\ref{eqt:SA}) together with (\ref{eqt:MMD}), 
both shared and domain specific features are exploited such that the two domains are well aligned geometrically and statistically.

\subsubsection{Overall Objective Function}
We formulate the JGSA method by incorporating the above five quantities ((\ref{eqt:target}), (\ref{eqt:between}), (\ref{eqt:within}), (\ref{eqt:MMD}), and (\ref{eqt:SA})) as follows:

\begin{footnotesize}
\begin{equation*}  
\max \frac{\mu\{\text{Target Var.}\}+\beta\{\text{Between Class Var.}\}}{\{\text{Distribution shift}\}+\lambda\{\text{Subspace shift}\}+\beta\{\text{Within Class Var.}\}}
\end{equation*}
\end{footnotesize}
where $\lambda,\mu,\beta$ are trade-off parameters to balance the importance of each quantity, and Var. indicates variance.

We follow~\cite{Ghifary2016} to further impose the constraint that $Tr(B^TB)$ is small to control the scale of $B$. Specifically, we aim at finding two coupled projections $A$ and $B$ by solving the following optimization function,

\begin{footnotesize}
\begin{align}
\begin{split}
\max_{A,B} \frac{Tr\bigg([\begin{matrix}
 A^T & B^T
 \end{matrix}] 
\Bigg[ \begin{matrix}
  \beta S_b & \mathbf{0} \\
   \mathbf{0} & \mu S_t
 \end{matrix}\Bigg] \Bigg[ \begin{matrix}
  A \\
  B
 \end{matrix} \Bigg]
\bigg)}{Tr\bigg([\begin{matrix}
 A^T &\hspace{-0.8em} B^T
 \end{matrix}] 
\Bigg[ \begin{matrix}
  M_s+\lambda I+\beta S_w &\hspace{-1em} M_{st}-\lambda I \\
  M_{ts}-\lambda I &\hspace{-1em} M_t+(\lambda+\mu) I
 \end{matrix}\Bigg]\hspace{-0.3em} \Bigg[ \begin{matrix}
  A \\
  B
 \end{matrix} \Bigg]
\bigg)}
\end{split}
\label{eqt:Obj}
\end{align}
\end{footnotesize}
where $I\in{\mathbb{R}^{d\times d}}$ is the identity matrix.

Minimizing the denominator of (\ref{eqt:Obj}) encourages small marginal and conditional distributions shifts, and small within class variance in the source domain. Maximizing the numerator of (\ref{eqt:Obj}) encourages large target domain variance, and large between class variance in the source domain. Similar to JDA, we also iteratively update the pseudo labels of target domain data using the learned transformations to improve the labelling quality until convergence.
\subsection{Optimization}
To optimize (\ref{eqt:Obj}), we rewrite $[\begin{matrix}
  A^T & B^T
 \end{matrix}]$ as $W^T$. Then the objective function and corresponding constraints can be rewritten as:
\begin{footnotesize}
 \begin{align}
\begin{split}
\max_{W} \frac{Tr\bigg(W^T 
\Bigg[ \begin{matrix}
  \beta S_b & \mathbf{0} \\
   \mathbf{0} & \mu S_t
 \end{matrix}\Bigg] W
\bigg)}{Tr\bigg(W^T 
\Bigg[ \begin{matrix}
  M_s+\lambda I+\beta S_w & M_{st}-\lambda I \\
  M_{ts}-\lambda I & M_t+(\lambda+\mu) I
 \end{matrix}\Bigg] W
\bigg)}
\end{split}
\label{eqt:Lag}
\end{align}
\end{footnotesize}
Note that the objective function is invariant to rescaling of $W$. Therefore, we rewrite objective function (\ref{eqt:Lag}) as 
\begin{footnotesize}
 \begin{align}
\begin{split}
 \max_{W} Tr\bigg(W^T 
\Bigg[ \begin{matrix}
  \beta S_b & \mathbf{0} \\
   \mathbf{0} & \mu S_t
 \end{matrix}\Bigg] W
\bigg)
\end{split}
\label{eqt:Lag2}
\end{align}
$$s.t. \quad Tr\bigg(W^T 
\Bigg[ \begin{matrix}
  M_s+\lambda I+\beta S_w & M_{st}-\lambda I \\
  M_{ts}-\lambda I & M_t+(\lambda+\mu) I
 \end{matrix}\Bigg] W
\bigg)=1$$
\end{footnotesize}
The Lagrange function of (\ref{eqt:Lag2}) is
\begin{footnotesize}
\begin{align}
\begin{split}
\hspace{-0.8em} L &=  Tr\bigg(W^T 
\Bigg[ \begin{matrix}
  \beta S_b & \mathbf{0} \\
   \mathbf{0} & \mu S_t
 \end{matrix}\Bigg] W
\bigg)\\
& + Tr\bigg(\bigg(W^T 
\Bigg[ \begin{matrix}
  M_s+\lambda I+\beta S_w & \hspace{-1em} M_{st}-\lambda I \\
  M_{ts}-\lambda I & \hspace{-1em}M_t+(\lambda+\mu) I
 \end{matrix}\Bigg] W - I
\bigg)\Phi \bigg)
\end{split}
\label{eqt:Lag3}
\end{align}
\end{footnotesize}
By setting the derivative $\frac{\partial L}{\partial W}=0$, we get:
\begin{footnotesize}
\begin{align}
\Bigg[ \begin{matrix}
  \beta S_b & \mathbf{0} \\
   \mathbf{0} & \mu S_t
 \end{matrix}\Bigg]W=
\Bigg[ \begin{matrix}
  M_s+\lambda I+\beta S_w &\hspace{-1em} M_{st}-\lambda I \\
  M_{ts}-\lambda I &\hspace{-1em} M_t+(\lambda+\mu) I
 \end{matrix}\Bigg] W \Phi 
 \label{eqt:final}
\end{align}
\end{footnotesize}
where $\Phi = diag(\lambda_1, ...,\lambda_k)$ are the k leading eigenvalues and $W = [W_1, ...,W_k]$ contains the corresponding eigenvectors, which can be solved analytically through generalized eigenvalue decomposition. Once the transformation matrix W is obtained, the subspaces A and B can be obtained easily. The pseudo code of JGSA is summarised in Algorithm~\ref{alg:GSA}.

\begin{algorithm}
\begin{small}
    \SetKwInOut{Input}{Input}
    \SetKwInOut{Output}{Output}

    \Input{Data and source labels: $X_s$, $X_t$, $Y_s$; Parameters: $\lambda=1$, $\mu=1$, $k$, $T$, $\beta$.}
    \Output{Transformation matrices: $A$ and $B$; Embeddings: $Z_s$, $Z_t$; Adaptive classifier: $f$.}
    Construct $S_t$, $S_b$, $S_w$, $M_s$, $M_t$, $M_{st}$, and $M_{ts}$ according to (\ref{eqt:St}), (\ref{eqt:between}), (\ref{eqt:within}), (\ref{eqt:Ms}), (\ref{eqt:Mt}), (\ref{eqt:Mst}), and (\ref{eqt:Mts}); Initialize pseudo labels in target domain $\hat{Y_t}$ using a classifier trained on original source domain data\;
    \Repeat{Convergence}{
      Solve the generalized eigendecompostion problem in Equation (\ref{eqt:final}) and select the k corresponding eigenvectors of k leading eigenvalues as the transformation $W$, and obtain subspaces $A$ and $B$\;
      Map the original data to respective subspaces to get the embeddings: $Z_s = A^TX_s$, $Z_t = B^TX_t$\;
      Train a classifier $f$ on $\{Z_s, Y_s\}$ to update pseudo labels in target domain $\hat{Y_t}=f(Z_t)$\;
      Update $M_s$, $M_t$, $M_{st}$, and $M_{ts}$ according to (\ref{eqt:Ms}), (\ref{eqt:Mt}), (\ref{eqt:Mst}), and (\ref{eqt:Mts}).
    }
    Obtain the final adaptive classifier $f$ on $\{Z_s, Y_s\}$.
    \caption{\small Joint Geometrical and Statistical Alignment}
    \label{alg:GSA}
\end{small}
\end{algorithm}

\subsection{Kernelization Analysis}
The JGSA method can be extended to nonlinear problems in a Reproducing Kernel Hilbert Space (RKHS) using some kernel functions $\phi$. 
We use the Representer Theorem $P = \Phi(X)A$ and $Q = \Phi(X)B$ to kernelize our method, where $X=[X_s, X_t]$ denotes all the source and target training samples, $\Phi(X)=[\phi(x_1),...,\phi(x_n)]$ and $n$ is the number of all samples. 
Hence, the objective function becomes,

\begin{footnotesize}
\begin{align}
\begin{split}
\max_{P,Q} \frac{Tr\bigg([\begin{matrix}
 P^T & Q^T
 \end{matrix}] 
\Bigg[ \begin{matrix}
  \beta S_b & \mathbf{0} \\
   \mathbf{0} & \mu S_t
 \end{matrix}\Bigg] \Bigg[ \begin{matrix}
  P \\
  Q
 \end{matrix} \Bigg]
\bigg)}{Tr\bigg([\begin{matrix}
 P^T & \hspace{-1em} Q^T
 \end{matrix}] 
\Bigg[ \begin{matrix}
  M_s+\lambda I+\beta S_w & \hspace{-1em} M_{st}-\lambda I \\
  M_{ts}-\lambda I & \hspace{-1em} M_t+(\lambda+\mu) I
 \end{matrix}\Bigg] \Bigg[ \begin{matrix}
  P \\
  Q
 \end{matrix} \Bigg]
\bigg)}
\end{split}
\label{eqt:Objker1}
\end{align}
\end{footnotesize}
where all the $X_t$'s are replaced by $\Phi(X_t)$ and all the $X_s$'s are replaced by $\Phi(X_s)$ in $S_t$, $S_w$, $S_b$, $M_s$, $M_t$, $M_{st}$, and $M_{ts}$ in the kernelized version.

We replace P and Q with $\Phi(X)A$ and $\Phi(X)B$ and obtain the objective function as follows,

\begin{footnotesize}
\begin{align}
\begin{split}
\hspace{-0.5em} \max_{A,B} \frac{Tr\bigg([\begin{matrix}
 A^T & B^T
 \end{matrix}] 
\Bigg[ \begin{matrix}
  \beta S_b & \mathbf{0} \\
   \mathbf{0} & \mu S_t
 \end{matrix}\Bigg] \Bigg[ \begin{matrix}
  A \\
  B
 \end{matrix} \Bigg]
\bigg)}{Tr\bigg([\begin{matrix}
 A^T & \hspace{-1em} B^T
 \end{matrix}] 
\Bigg[ \begin{matrix}
  M_s+\lambda K+\beta S_w & \hspace{-1em} M_{st}-\lambda K \\
  M_{ts}-\lambda K & \hspace{-1em} M_t+(\lambda+\mu) K
 \end{matrix}\Bigg]\hspace{-0.3em} \Bigg[ \begin{matrix}
  A \\
  B
 \end{matrix} \Bigg]
\bigg)}
\end{split}
\label{eqt:Objker2}
\end{align}
\end{footnotesize}
where $S_t = \tilde{K_t}\tilde{K_t}^T$, $S_w = K_sH_s^{(c)}K_s^T$, with $K = \Phi(X)^T\Phi(X)$, $K_s = \Phi(X)^T\Phi(X_s)$, $K_t = \Phi(X)^T\Phi(X_t)$, $\tilde{K_t} = K_t-\mathbf{1}_{t}K-K_t\mathbf{1}_{n}+\mathbf{1}_{t}K\mathbf{1}_{n}$, $\mathbf{1}_t\in{\mathbb{R}^{n_t\times n}}$ and $\mathbf{1}_n\in{\mathbb{R}^{n\times n}}$ are matrices with all $\frac{1}{n}$.
In $S_b$, $m_s^{(c)}=\frac{1}{n_s^{(c)}}\sum_{i=1}^{n_s^{(c)}}k_i^{(c)}$, $\bar{m}_s = \frac{1}{n_s}\sum_{i=1}^{n_s}k_i$, with $k_i = \Phi(X)^T\phi(x_i)$. 
In MMD terms,
$M_s = K_s(L_s+\sum_{c=1}^C L_s^{(c)})K_s^T$,
$M_t = K_t(L_t+\sum_{c=1}^C L_t^{(c)})K_t^T$, 
$M_{st} = K_s(L_{st}+\sum_{c=1}^C L_{st}^{(c)})K_t^T $,
$M_{ts} = K_t(L_{ts}+\sum_{c=1}^C L_{ts}^{(c)})K_s^T $.
Once the kernelized objective function (\ref{eqt:Objker2}) is obtained, we can simply solve it in the same way as the original objective function to compute $A$ and $B$. 

\section{Experiments}
In this section, we first conduct experiments on a synthetic dataset to verify the effectiveness of the JGSA methods. Then we evaluate our method for cross-domain object recognition, cross-domain digit recognition, and cross dataset RGB-D-based action recognition. The codes are available online\footnote{\url{http://www.uow.edu.au/~jz960/}}. We compare our method with several state-of-the-art methods: subspace alignment (SA)~\cite{Fernando2013}, subspace distribution alignment (SDA)~\cite{Sun2015}, geodesic flow kernel (GFK)~\cite{Gong2012}, transfer component analysis (TCA)~\cite{Pan2011}, joint distribution analysis (JDA)~\cite{Long2013}, transfer joint matching (TJM)~\cite{Long2014}, 
scatter component analysis (SCA)~\cite{Ghifary2016}, optimal transport (OTGL)~\cite{Courty2016}, and kernel manifold alignment (KEMA)~\cite{Tuia2016}. We use the parameters recommended by the original papers for all the baseline methods. For JGSA, we fix $\lambda=1$, $\mu=1$ in all the experiments, such that the distribution shift, subspace shift, and target variance are treated as equally important. We empirically verified that the fixed parameters can obtained promising results on different types of tasks. Hence, the subspace dimension $k$, number of iteration $T$, and regularization parameter $\beta$ are free parameters.
\subsection{Synthetic Data}
\label{sec:synthetic}
Here, we aim to synthesize samples of data to demonstrate that our method can keep the domain structures as well as reduce the domain shift. The synthesized source and target domain samples are both draw from a mixture of three RBFian distributions. Each RBFian distribution represents one class. The global means, as well as the means of the third class are shifted between domains. The original data are 3-dimensional. We set the dimensionality of the subspaces to 2 for all the methods. 

\begin{figure*}[ht!]
    \centering
    \begin{subfigure}[t]{0.245\textwidth}
        \centering
        \includegraphics[scale=0.2]{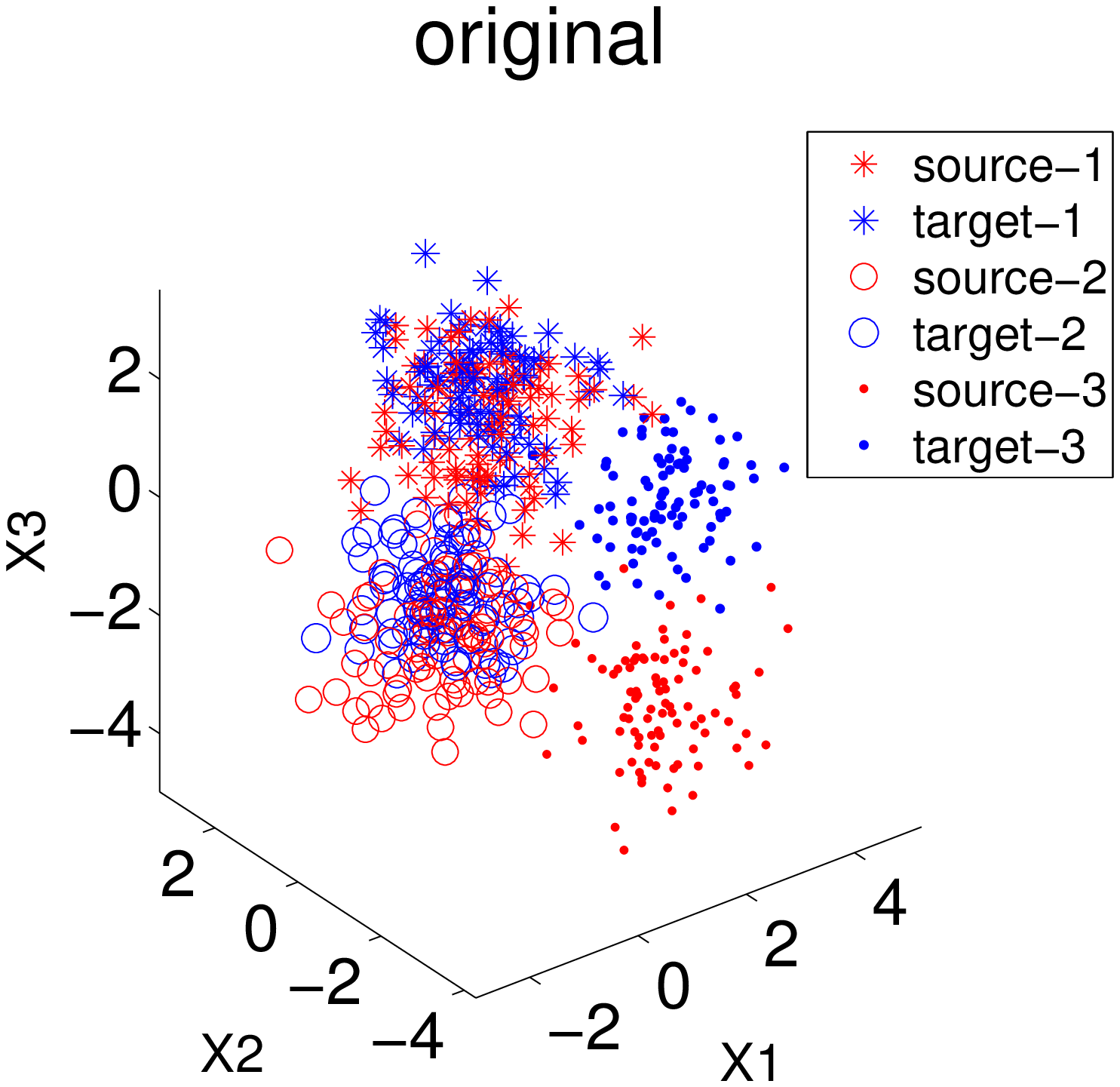}
\label{fig:DA1}
    \end{subfigure}%
~
	\begin{subfigure}[t]{0.235\textwidth}
        \centering
        \includegraphics[scale=0.2]{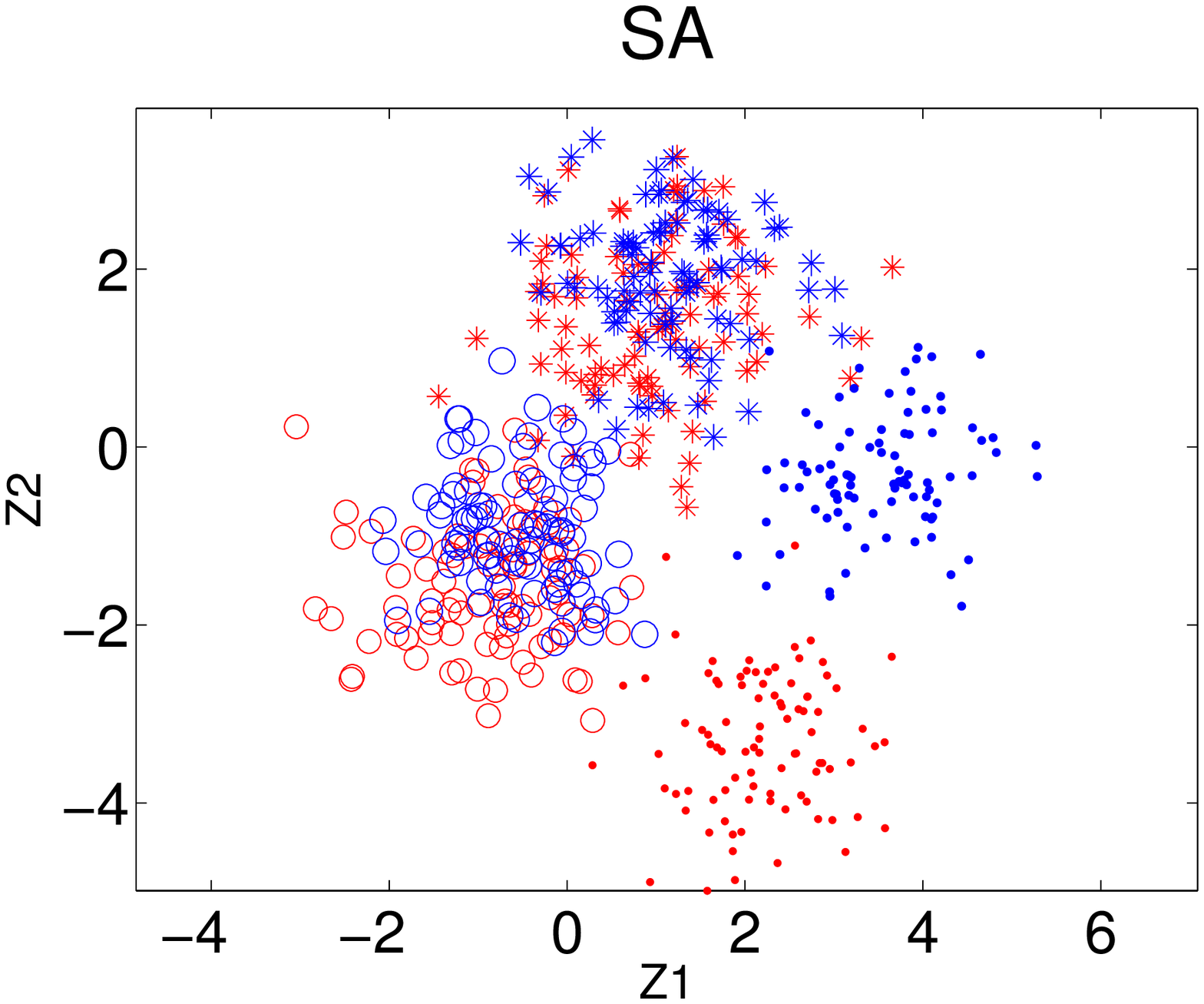}
\label{fig:DA2}
    \end{subfigure}
~
    \begin{subfigure}[t]{0.235\textwidth}
        \centering
        \includegraphics[scale=0.2]{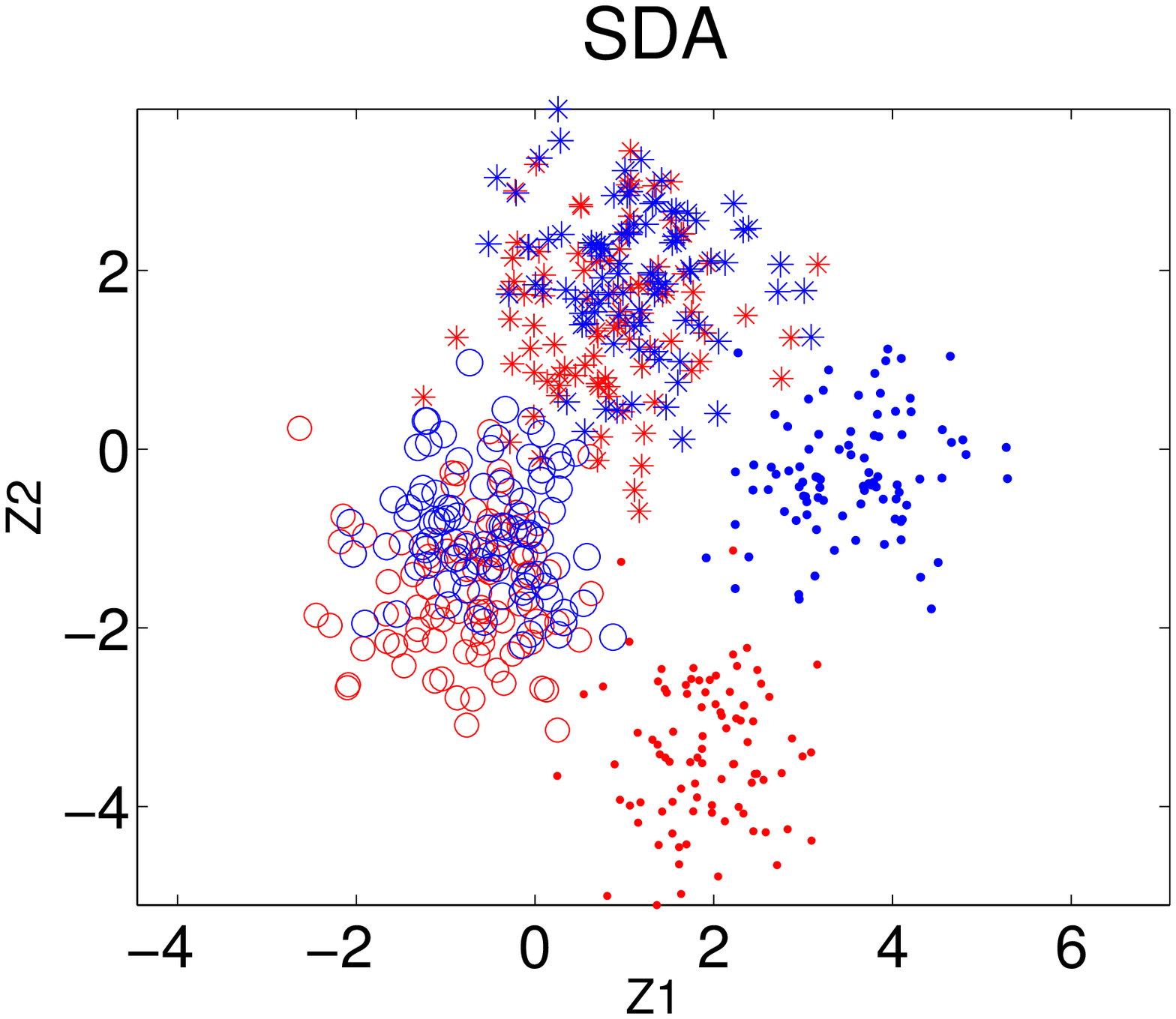}
\label{fig:DA3}
    \end{subfigure}
~
    \begin{subfigure}[t]{0.235\textwidth}
        \centering
        \includegraphics[scale=0.2]{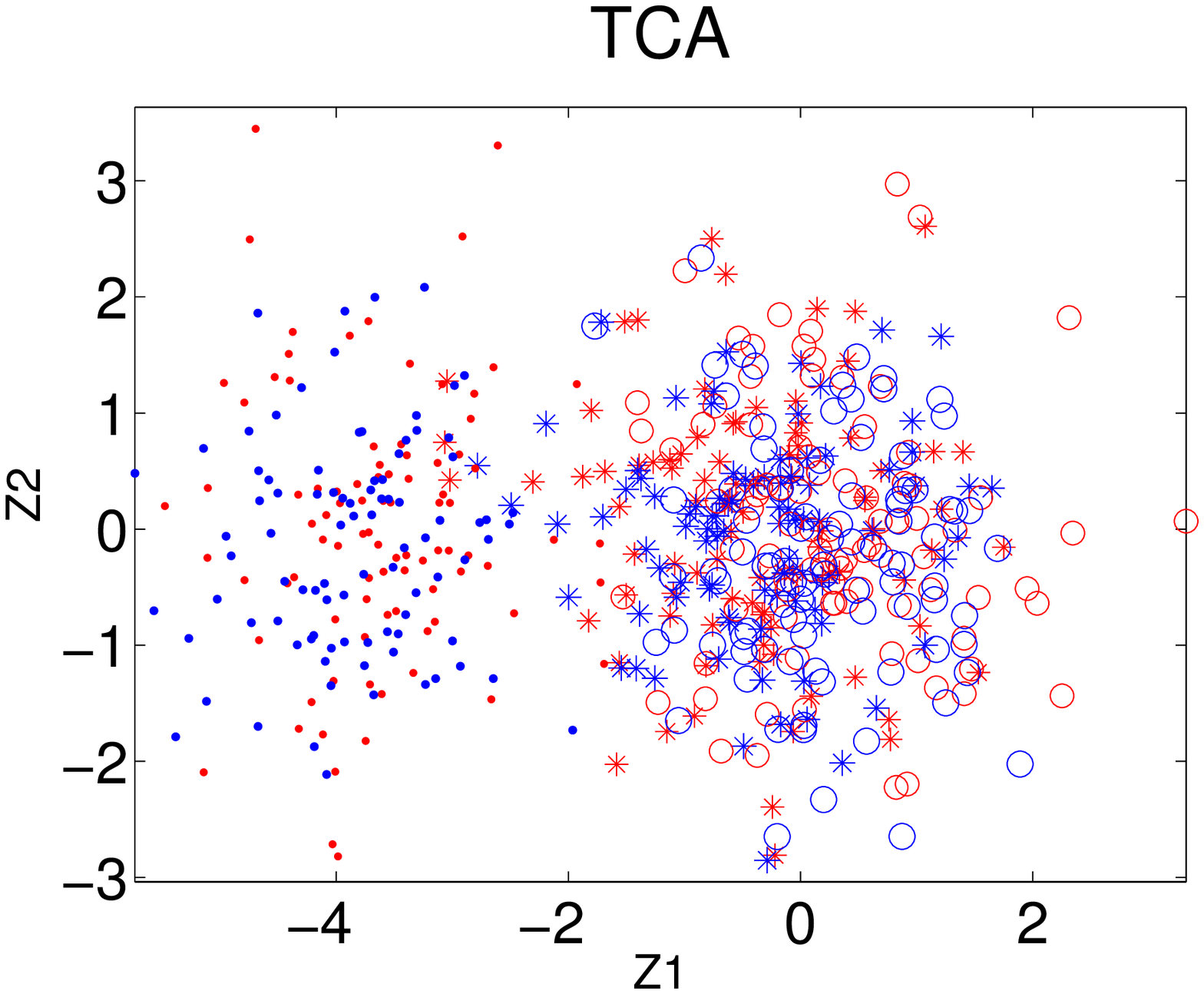}
\label{fig:DA4}
    \end{subfigure}\\
\vspace{-1em}
    \begin{subfigure}[t]{0.235\textwidth}
        \centering
        \includegraphics[scale=0.2]{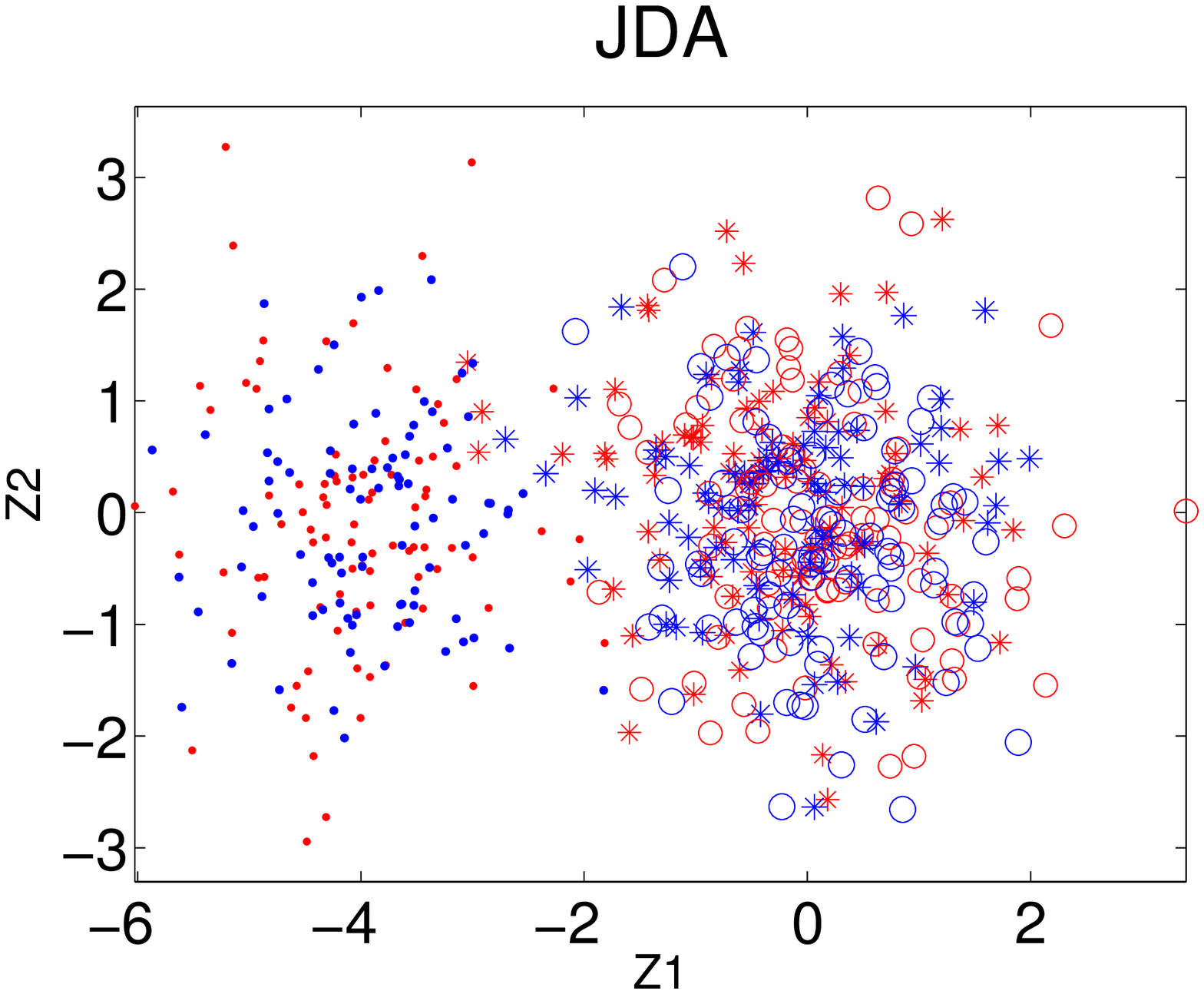}
\label{fig:DA5}
    \end{subfigure}
~
    \begin{subfigure}[t]{0.235\textwidth}
        \centering
        \includegraphics[scale=0.2]{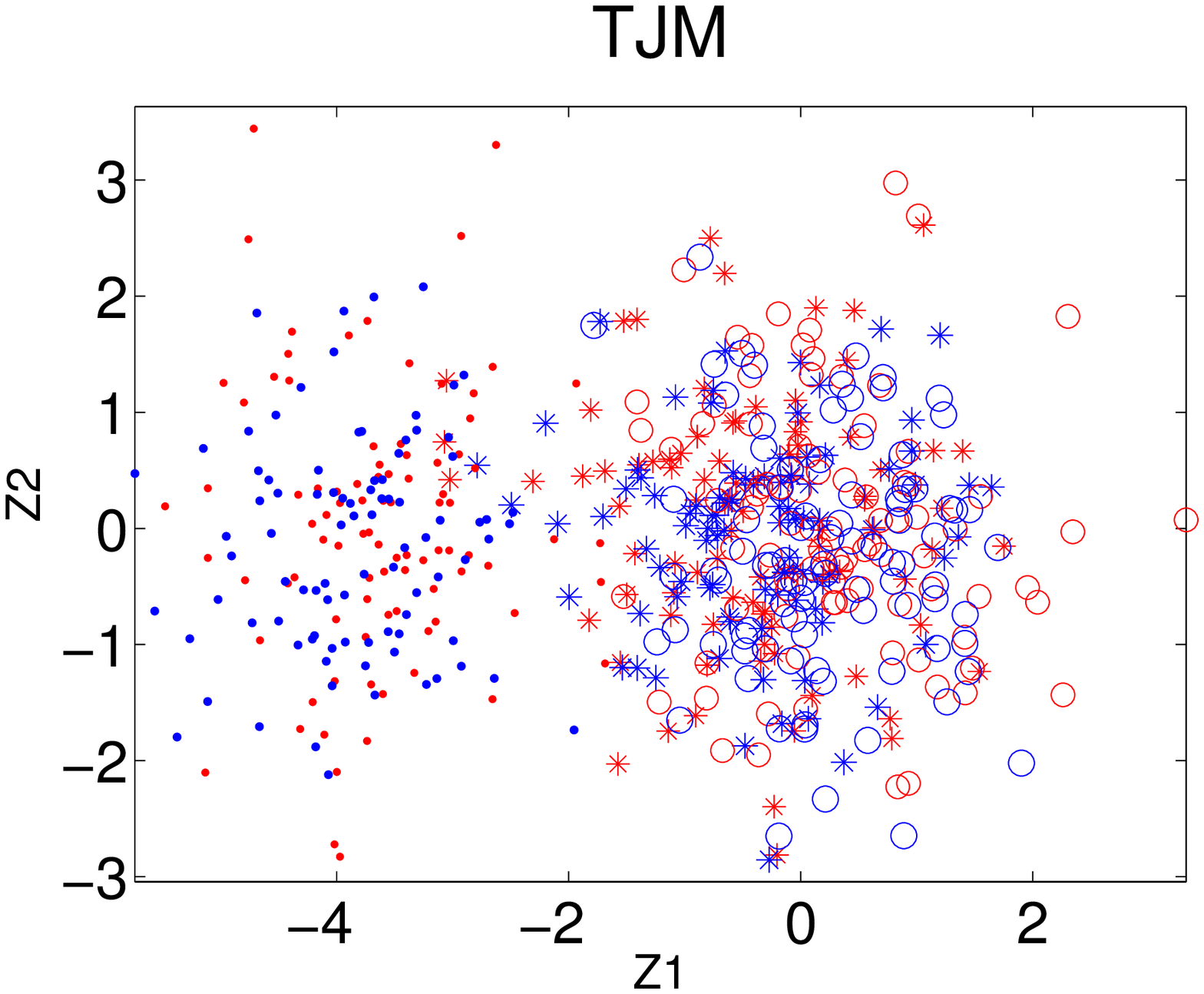}
\label{fig:DA6}
    \end{subfigure}
~
    \begin{subfigure}[t]{0.235\textwidth}
        \centering
        \includegraphics[scale=0.2]{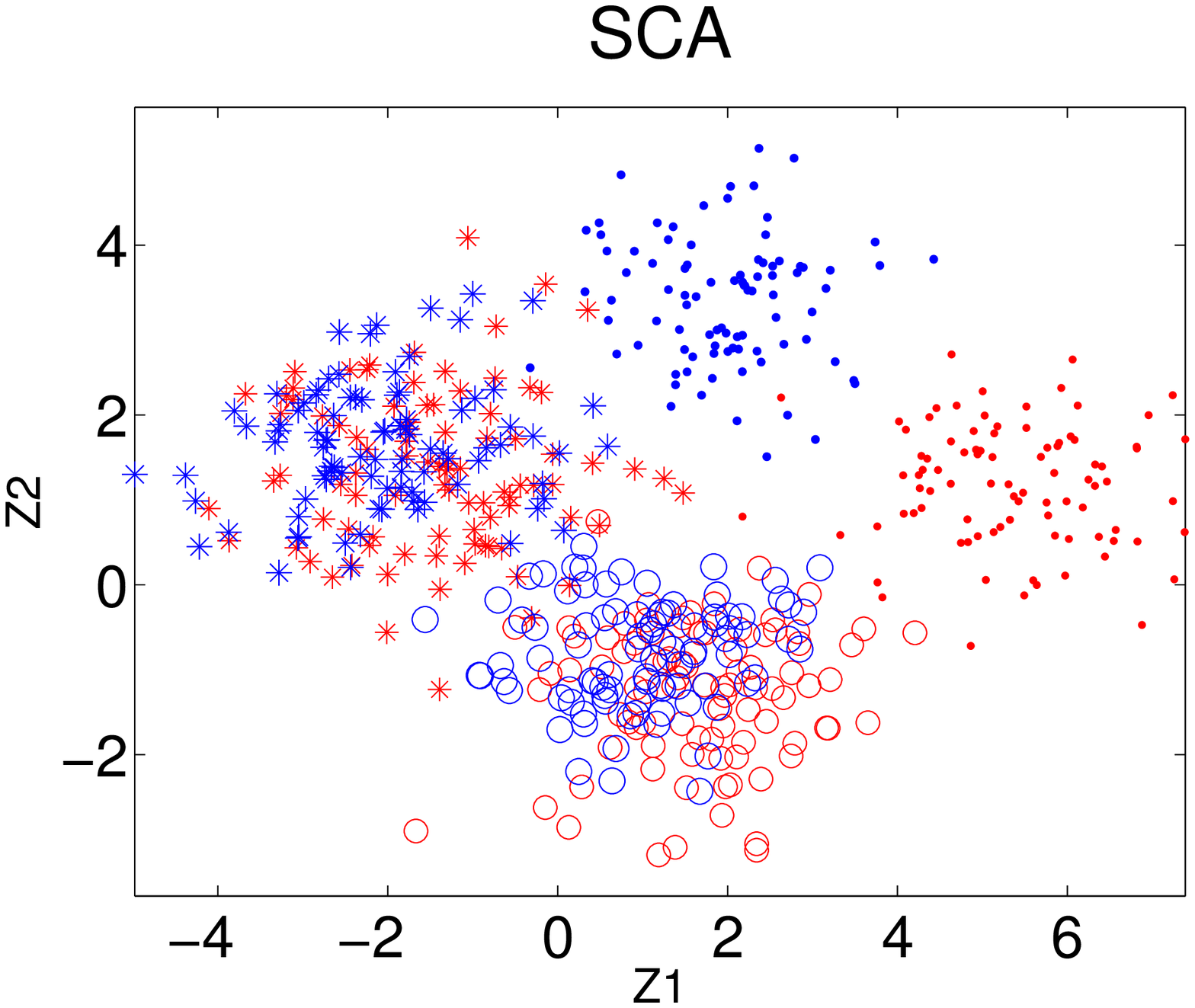}
\label{fig:DA7}
    \end{subfigure}
~
    \begin{subfigure}[t]{0.235\textwidth}
        \centering
        \includegraphics[scale=0.2]{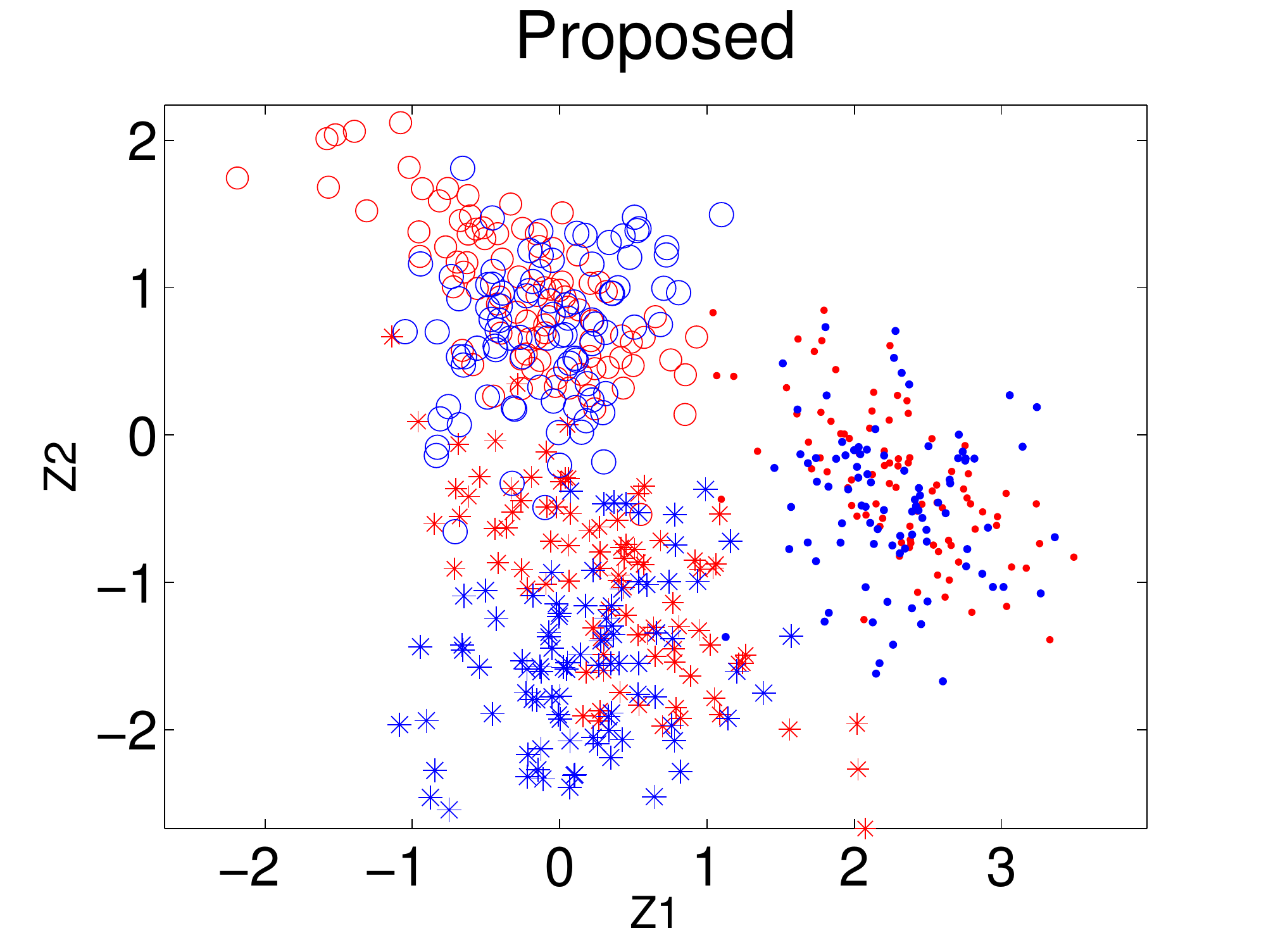}
\label{fig:DA8}
    \end{subfigure}
    \caption{Comparisons of baseline domain adaptation methods and the proposed JGSA method on the synthetic data }
    \label{fig:toy}
\end{figure*}

Figure~\ref{fig:toy} illustrates the original synthetic dataset and domain adaptation results of different methods on the dataset. It can be seen that after SA method the divergences between domains are still large after aligning the subspaces. Hence, the aligned subspaces are not optimal for reduce the domain shift if the distribution divergence is not considered. The SDA method does not demonstrate obvious improvement over SA, since the variance shift is reduced based upon the aligned subspaces (which may not be optimal) as in SA. TCA method reduces the domain shift effectively. However, two of the classes are mixed up since there may not exist a unified subspace to reduce domain shift and preserve the original information simultaneously. Even with conditional distribution shift reduction (JDA) or instances reweighting (TJM), the class-1 and class-2 still cannot be distinguished. SCA considers the total scatter, domain scatter, and class scatter using a unified mapping. However, there may not exist such a common subspace that satisfies all the constraints.
Obviously, JGSA aligns the two domains well even though the shift between source and target domains is large.

\subsection{Real World Datasets}
\begin{figure}[ht!]
\begin{tabular}{cccc}

\begin{minipage}{.09\textwidth}
\includegraphics[scale=0.1]{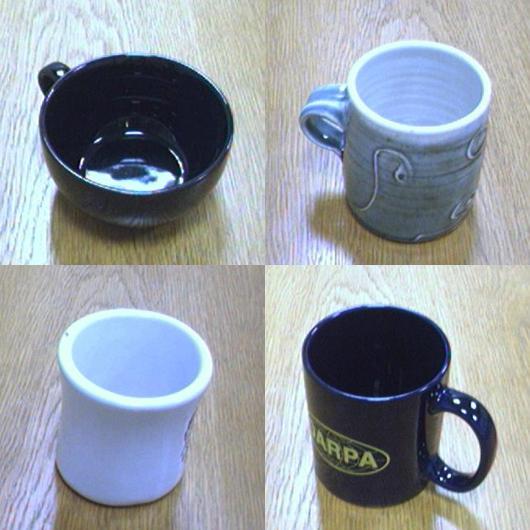}
\end{minipage} & 
\begin{minipage}{.09\textwidth}
\includegraphics[scale=0.1]{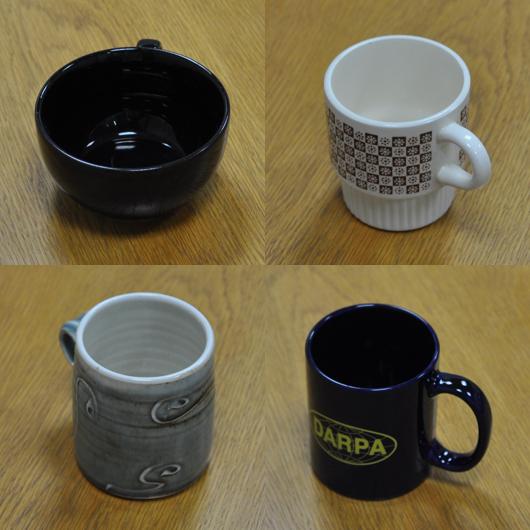}
\end{minipage} &
\begin{minipage}{.09\textwidth}
\includegraphics[scale=0.1]{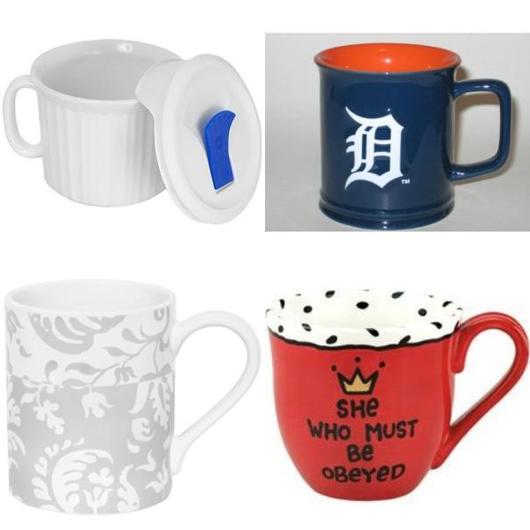}
\end{minipage} & 
\begin{minipage}{.09\textwidth}
\includegraphics[scale=0.1]{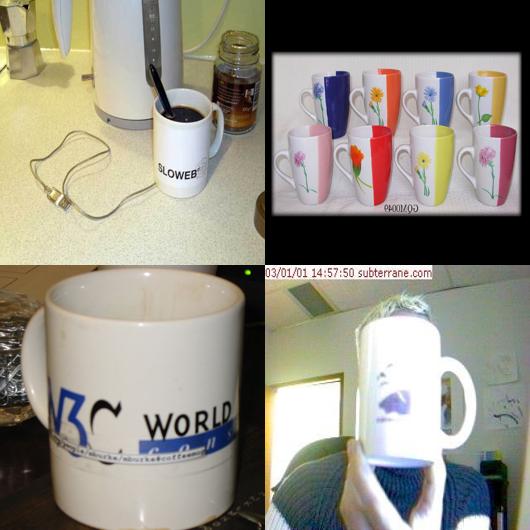}
\end{minipage} \\
Webcam & DSLR & Amazon & Caltech\\
\end{tabular}
\begin{tabular}{cc}
\begin{minipage}{.2\textwidth}
\includegraphics[scale=0.68]{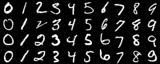}
\end{minipage} & 
\begin{minipage}{.2\textwidth}
\includegraphics[scale=0.68]{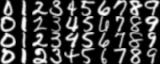}
\end{minipage} \\
MNIST & USPS\\
\end{tabular}
\begin{tabular}{ccc}

\begin{minipage}{.13\textwidth}
\includegraphics[scale=0.22]{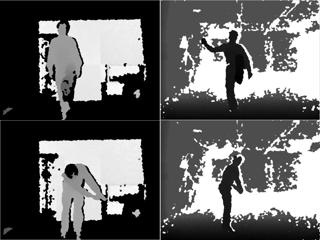}
\end{minipage} & 
\begin{minipage}{.13\textwidth}
\includegraphics[scale=0.22]{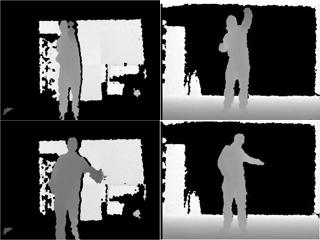}
\end{minipage} &
\begin{minipage}{.13\textwidth}
\includegraphics[scale=0.22]{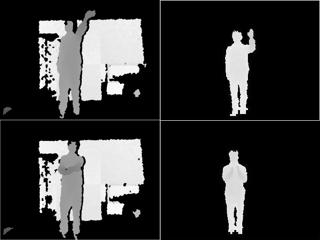}
\end{minipage} \\
MSR vs. G3D & MSR vs. MAD  & MSR vs. UTD 
\end{tabular}
\caption{Sample images of object datasets, digit datasets, and sample video frames of depth map of RGB-D-based action datasets.}
\label{fig:datasets}
\vspace{-1em}
\end{figure}

We evaluate our method on three cross-domain visual recognition tasks: object recognition (Office, Caltech-256), hand-written digit recognition (USPS, MNIST), and RGB-D-based action recognition (MSRAction3DExt, G3D, UTD-MHAD, and MAD). The sample images or video frames are shown in Figure~\ref{fig:datasets}. 

\subsubsection{Setup}
\label{sec:setup}
\paragraph{Object Recognition} We adopt the public Office+Caltech object datasets released by Gong et al.~\cite{Gong2012}. This dataset contains images from four different domains: Amazon (images downloaded from online merchants), Webcam (low-resolution images by a web camera), DSLR (high-resolution images by a digital SLR camera), and Caltech-256. Amazon, Webcam, and DSLR are three datasets studied in~\cite{Saenko2010} for the effects of domain shift. Caltech-256~\cite{Griffin2007} contains 256 object classes downloaded from Google images. Ten classes common to four datasets are selected: \textit{backpack, bike, calculator, head-phones, keyboard, laptop, monitor, mouse, mug, and projector}.
Two types of features are considered: SURF descriptors (which are encoded with 800-bin histograms with the codebook trained from a subset of Amazon images), and $Decaf_6$ features (which are the activations of the $6th$ fully connected layer of a convolutional network trained on imageNet).
As suggested by~\cite{Gong2012}, 1-Nearest Neighbor Classifier (NN) is chosen as the base classifier. For the free parameters, we set $k=30$, $T=10$, and $\beta=0.1$.

\paragraph{Digit Recognition}
For cross-domain hand-written digit recognition task, we use MNIST~\cite{LeCun1998} and USPS~\cite{Hull1994} datasets to evaluate our method. MNIST dataset contains a training set of 60,000 examples, and a test set of 10,000 examples of size 28$\times$28. USPS dataset consists of 7,291 training images and 2,007 test images of size 16$\times$16. Ten shared classes of the two datasets are selected. We follow the settings of~\cite{Long2013,Long2014} to construct a pair of cross-domain datasets USPS $\rightarrow$ MNIST by randomly sampling 1,800 images in USPS to form the source data, and randomly sampling 2,000 images in MNIST to form the target data. Then source and target pair are switched to form another dataset MNIST $\rightarrow$ USPS. All images are uniformly rescaled to size 16$\times$16, and each image is represented by a feature vector encoding the gray-scale pixel values.
For the free parameters, we set $k=100$, $T=10$, and $\beta=0.01$.

\begin{table*}[ht!]
\caption{Accuracy(\%) on cross-domain object datasets. Notation for datasets: Caltech:C; Amazon:A; Webcam:W; DSLR:D.}
\label{tab:object}
\vspace{-1em}
\begin{center}
\begin{scriptsize}
\begin{tabu}{ | m{0.7cm} | m{0.5cm} | m{0.5cm} | m{0.5cm} | m{0.5cm} | m{0.5cm} | m{0.5cm} | m{0.5cm} | m{0.5cm} | m{0.7cm} | m{0.7cm} | m{0.7cm} |[1.5pt] m{0.5cm} | m{0.7cm} | m{0.7cm} | m{0.7cm}| m{0.7cm} |}
\hline
Feature & \multicolumn{11}{c|[1.5pt]}{SURF} & \multicolumn{5}{c|}{$Decaf_6$}\\
\hline
	data & Raw & SA & SDA & GFK & TCA & JDA & TJM & SCA & JGSA primal & JGSA linear & JGSA RBF & JDA & OTGL & JGSA primal & JGSA linear & JGSA RBF \\ \hline \hline
	C$\rightarrow$A & 36.01 & 49.27 & 49.69 & 46.03 & 45.82 & 45.62 & 46.76 & 45.62 & 51.46 & 52.30 & \textbf{53.13}
 & 90.19 & \textbf{92.15} & 91.44 & 91.75 & 91.13\\ \hline
	C$\rightarrow$W & 29.15 & 40.00 & 38.98 & 36.95 & 31.19 & 41.69 & 38.98 & 40.00 & 45.42 & 45.76 & \textbf{48.47}
 & 85.42 & 84.17 & \textbf{86.78} & 85.08 & 83.39\\ \hline
	C$\rightarrow$D & 38.22 & 39.49 & 40.13 & 40.76 & 34.39 & 45.22 & 44.59 & 47.13 & 45.86 & \textbf{48.41} & \textbf{48.41} & 85.99 & 87.25 & \textbf{93.63} & 92.36 & 92.36\\ \hline
	A$\rightarrow$C & 34.19 & 39.98 & 39.54 & 40.69 & \textbf{42.39} & 39.36 & 39.45 & 39.72 & 41.50 & 38.11 & 41.50 & 81.92 & \textbf{85.51} & 84.86 & 85.04 & 84.86\\ \hline
	A$\rightarrow$W & 31.19 & 33.22 & 30.85 & 36.95& 36.27 & 37.97 & 42.03 & 34.92 & 45.76 & \textbf{49.49} & 45.08
 & 80.68 & 83.05 & 81.02 & \textbf{84.75} & 80.00 \\ \hline
	A$\rightarrow$D & 35.67 & 33.76 & 33.76 & 40.13 & 33.76 & 39.49 & 45.22 & 	39.49 & \textbf{47.13} & 45.86 & 45.22 & 81.53 & 85.00 & \textbf{88.54} & 85.35 & 84.71\\ \hline
	W$\rightarrow$C & 28.76 & \textbf{35.17} & 34.73 & 24.76 & 29.39 & 31.17 & 30.19 & 31.08 & 33.21 & 32.68 & 33.57 & 81.21 & 81.45 & \textbf{84.95} & 84.68 & 84.51\\ \hline
	W$\rightarrow$A & 31.63 & 39.25 & 39.25 & 27.56 & 28.91 & 32.78 & 29.96 & 
 29.96 & 39.87 & \textbf{41.02} & 40.81 & 90.71 & 90.62 & 90.71 & \textbf{91.44} & 91.34\\ \hline
	W$\rightarrow$D & 84.71 & 75.16 & 75.80 & 85.35 & 89.17 & 89.17 & 89.17 & 87.26 & \textbf{90.45} & \textbf{90.45} & 88.54
 & \textbf{100} & 96.25 & \textbf{100} & \textbf{100} & \textbf{100} \\ \hline
	D$\rightarrow$C & 29.56 & 34.55 & \textbf{35.89} & 29.30 & 30.72 & 31.52 & 31.43 & 30.72 & 29.92 & 30.19 & 30.28 & 80.32 & 84.11 & \textbf{86.20} & 85.75 & 84.77\\ \hline
	D$\rightarrow$A & 28.29 & \textbf{39.87} & 38.73 & 28.71 & 31.00 & 33.09 & 32.78 & 31.63 & 38.00 & 36.01 & 38.73 & 91.96 & \textbf{92.31} & 91.96 & 92.28 & 91.96\\ \hline
	D$\rightarrow$W & 83.73 & 76.95 & 76.95 & 80.34 & 86.10 & 89.49 & 85.42 & 84.41 & 91.86 & 91.86 & \textbf{93.22}
& 99.32 & 96.29 & \textbf{99.66} & 98.64 & 98.64\\ \hline
	Average & 40.93 & 44.72 & 44.52 & 43.13 & 43.26 & 46.38 & 46.33 & 45.16 & 50.04 & 50.18 & \textbf{50.58} & 87.44 & 88.18 & \textbf{89.98}  & 89.76 & 88.97\\ \hline
\end{tabu}
\end{scriptsize}
\end{center}
\end{table*}

\begin{table*}[ht!]
\caption{Accuracy (\%) on cross-domain digit datasets.}
\label{tab:digit}
\vspace{-1em}
\begin{center}
\begin{small}
\begin{tabular}{ | l | l | l | l | l | l | l | l | l | l | l | }
\hline
	data & Raw & SA & SDA & GFK & TCA & JDA & TJM & SCA & JGSA primal \\
\hline \hline
	MNIST$\rightarrow$USPS & 65.94 & 67.78 & 65.00 & 61.22 & 56.33 & 67.28 & 63.28 & 65.11 & \textbf{80.44} \\ \hline
	USPS$\rightarrow$MNIST & 44.70 & 48.80 & 35.70 & 46.45 & 51.20 & 59.65 & 52.25 & 48.00 & \textbf{68.15} \\ \hline
	Average & 55.32 & 58.29 & 50.35 & 56.84 & 53.77 & 63.47 & 57.77 & 56.56 & \textbf{74.30}  \\ \hline
\end{tabular}
\end{small}
\end{center}
\end{table*}

\begin{table*}[ht!]
\begin{center}
\caption{Accuracy (\%) on cross-dataset RGB-D-based action datasets.}
\label{tab:action}
\begin{small}
\begin{tabular}{ | l | l | l | l | l | l | l | l | l | l | l | l | l | }
\hline
	data & Raw & SA & SDA & TCA & JDA & TJM & SCA & JGSA linear \\\hline \hline
	MSR$\rightarrow$G3D & 72.92 & 77.08 & 73.96 & 68.75 & 82.29 & 70.83 & 70.83 & \textbf{89.58} \\\hline
	G3D$\rightarrow$MSR & 54.47 & \textbf{68.09} & 67.32 & 50.58 & 65.37 & 63.04 & 55.25 & 66.93 \\\hline
	MSR$\rightarrow$UTD & 66.88 & 73.75 & 73.75 & 65.00 & \textbf{77.50} & 65.00 & 64.38 & 76.88 \\\hline
	UTD$\rightarrow$MSR & 62.93 & \textbf{67.91} & 66.67 & 57.63 & 61.06 & 60.12 & 55.14 & 61.37 \\\hline
	MSR$\rightarrow$MAD & 80.71 & 85.00 & 83.57 & 79.29 & 82.86 & 82.14 & 78.57 & \textbf{86.43} \\\hline
	MAD$\rightarrow$MSR & 80.09 & 81.48 & 80.56 & 81.02 & 83.33 & 79.63 & 79.63 & \textbf{85.65} \\\hline
	Average & 69.67 & 75.55 & 74.30 & 67.05 & 75.40 & 70.13 & 67.30 & \textbf{77.81} \\\hline

\end{tabular}
\end{small}
\end{center}
\end{table*}

\begin{figure*}[ht!]
    \begin{subfigure}[t]{0.3\textwidth}
\pgfplotsset{grid style={dotted,very thin,lightgray},compat=1.5.1}
\resizebox{150pt}{120pt}{%
\begin{tikzpicture}
\begin{axis}[
xtick={-15,-13,-11,-9,-7,-5,-3,-1,1},
ytick={10,20,30,40,50,60,70,80,90,100},
ymin=15,ymax=91,
yticklabel style={/pgf/number format/precision=0},
xticklabels={$2^{-15}$,$2^{-13}$,$2^{-11}$,$2^{-9}$,$2^{-7}$,$2^{-5}$,$2^{-3}$,$2^{-1}$,$2^{1}$},
  xlabel=$\beta$ value,
  ylabel=Accuracy(\%),
  grid=both,
  legend pos= south west,
  legend columns=2,
  legend style={font=\fontsize{8}{8}\selectfont,/tikz/column 2/.style={
                column sep=5pt,
            },}]


\addplot+[line width=0.25mm,solid,color=black,mark=square*,mark options={color=black}] table [x=beta, y=MSR->MAD]{data.dat};
\addlegendentry{MSR$\rightarrow$MAD}

\addplot+[line width=0.25mm,dashed,color =black,mark=square*,mark options={color=black}] table [x=beta, y=MSR->MADbaseline]{data.dat};
\addlegendentry{MSR$\rightarrow$MADbaseline}

\addplot+[line width=0.25mm,color =blue,mark=diamond*,mark options={color=blue}] table [x=beta, y=USPS->MNIST]{data.dat};
\addlegendentry{USPS$\rightarrow$MNIST}

\addplot+[line width=0.25mm,dashed,color =blue,mark=diamond*,mark options={color=blue}] table [x=beta, y=USPS->MNISTbaseline]{data.dat};
\addlegendentry{USPS$\rightarrow$MNISTbaseline}

\addplot+[line width=0.25mm,color=red,mark=*,mark options={color=red}] table [x=beta, y=W->A]{data.dat};
\addlegendentry{W$\rightarrow$A}

\addplot+[line width=0.25mm,dashed,color =red,mark=*,mark options={color=red}] table [x=beta, y=W->Abaseline]{data.dat};
\addlegendentry{W$\rightarrow$Abaseline}

\end{axis}
\end{tikzpicture}
}
\vspace{-2em}
\caption{regularization parameter $\beta$} \label{fig:beta}
    \end{subfigure}%
~
    \begin{subfigure}[t]{0.3\textwidth}

\pgfplotsset{grid style={dotted,very thin,lightgray},compat=1.5.1}
\resizebox{150pt}{120pt}{%
\begin{tikzpicture}
\begin{axis}[
xtick={10,30,50,70,90,110,130,150,170,190},
ytick={10,20,30,40,50,60,70,80,90,100},
ymin=15,ymax=91,
yticklabel style={/pgf/number format/precision=0},
  xlabel=$k$ value,
  ylabel=Accuracy(\%),
  grid=both,  legend pos= south west,
  legend columns=2,
  legend style={font=\fontsize{8}{8}\selectfont,/tikz/column 2/.style={
                column sep=5pt,
            },}]


\addplot+[line width=0.25mm,solid,color=black,mark=square*,mark options={color=black}] table [x=k, y=MSR->MAD]{datak.dat};
\addlegendentry{MSR$\rightarrow$MAD}

\addplot+[line width=0.25mm,dashed,color =black,mark=square*,mark options={color=black}] table [x=k, y=MSR->MADbaseline]{datak.dat};
\addlegendentry{MSR$\rightarrow$MADbaseline}

\addplot+[line width=0.25mm,color =blue,mark=diamond*,mark options={color=blue}] table [x=k, y=USPS->MNIST]{datak.dat};
\addlegendentry{USPS$\rightarrow$MNIST}

\addplot+[line width=0.25mm,dashed,color =blue,mark=diamond*,mark options={color=blue}] table [x=k, y=USPS->MNISTbaseline]{datak.dat};
\addlegendentry{USPS$\rightarrow$MNISTbaseline}

\addplot+[line width=0.25mm,color=red,mark=*,mark options={color=red}] table [x=k, y=W->A]{datak.dat};
\addlegendentry{W$\rightarrow$A}

\addplot+[line width=0.25mm,dashed,color =red,mark=*,mark options={color=red}] table [x=k, y=W->Abaseline]{datak.dat};
\addlegendentry{W$\rightarrow$Abaseline}

\end{axis}
\end{tikzpicture}
}
\vspace{-2em}
\caption{dimentionality of subspace $k$} \label{fig:k}
    \end{subfigure}%
~
    \begin{subfigure}[t]{0.3\textwidth}

\pgfplotsset{grid style={dotted,very thin,lightgray},compat=1.5.1}
\resizebox{150pt}{120pt}{%
\begin{tikzpicture}
\begin{axis}[
xtick={1,2,3,4,5,6,7,8,9,10},
ytick={10,20,30,40,50,60,70,80,90,100},
ymin=15,ymax=91,
yticklabel style={/pgf/number format/precision=0},
  xlabel=$T$ value,
  ylabel=Accuracy(\%),
  grid=both,  legend pos= south west,
  legend columns=2,
  legend style={font=\fontsize{8}{8}\selectfont,/tikz/column 2/.style={
                column sep=5pt,
            },}]


\addplot+[line width=0.25mm,solid,color=black,mark=square*,mark options={color=black}] table [x=T, y=MSR->MAD]{datat.dat};
\addlegendentry{MSR$\rightarrow$MAD}

\addplot+[line width=0.25mm,dashed,color =black,mark=square*,mark options={color=black}] table [x=T, y=MSR->MADbaseline]{datat.dat};
\addlegendentry{MSR$\rightarrow$MADbaseline}

\addplot+[line width=0.25mm,color =blue,mark=diamond*,mark options={color=blue}] table [x=T, y=USPS->MNIST]{datat.dat};
\addlegendentry{USPS$\rightarrow$MNIST}

\addplot+[line width=0.25mm,dashed,color =blue,mark=diamond*,mark options={color=blue}] table [x=T, y=USPS->MNISTbaseline]{datat.dat};
\addlegendentry{USPS$\rightarrow$MNISTbaseline}

\addplot+[line width=0.25mm,color=red,mark=*,mark options={color=red}] table [x=T, y=W->A]{datat.dat};
\addlegendentry{W$\rightarrow$A}

\addplot+[line width=0.25mm,dashed,color =red,mark=*,mark options={color=red}] table [x=T, y=W->Abaseline]{datat.dat};
\addlegendentry{W$\rightarrow$Abaseline}

\end{axis}
\end{tikzpicture}
}
\vspace{-2em}
\caption{number of iteration $T$} \label{fig:T}
    \end{subfigure}%
\vspace{-0.5em}
\caption{Parameter sensitivity study of JGSA on different types of datasets}
\vspace{-1em}
\label{fig:params}
\end{figure*}
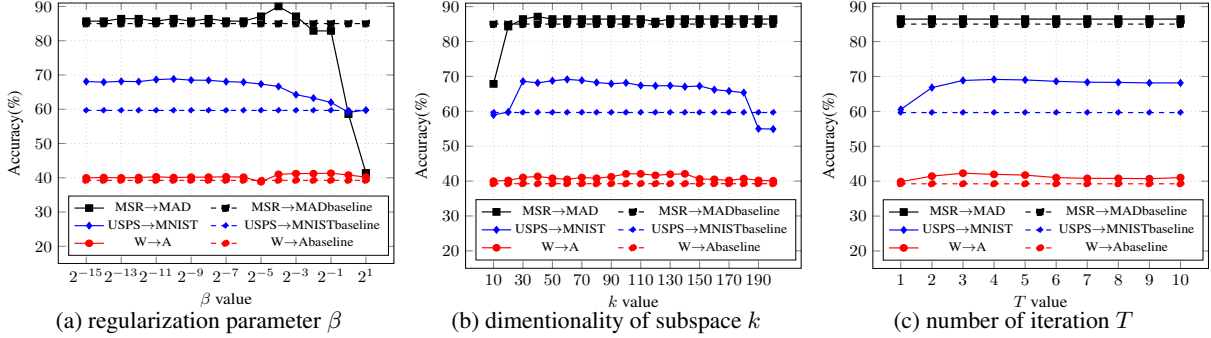

\vspace{-1em}
\paragraph{RGB-D-based Action Recognition}
For cross-dataset RGB-D-based Action Recognition, four RGB-D-based Action Recognition datasets are selected, namely MSRAction3DExt~\cite{Li2010,Wang2016}, UTD-MHAD~\cite{Chen2015b}, G3D\cite{Bloom2012}, and MAD~\cite{Huang2014}. All the four datasets are captured by both RGB and depth sensors. We select the shared actions between MSRAction3DExt and other three datasets to form 6 dataset pairs. There are 8 common actions between MSRAction3DExt and G3D: \textit{wave, forward punch, hand clap, forward kick, jogging, tennis swing, tennis serve,} and \textit{golf swing}. There are 10 common actions between MSRAction3DExt and UTD-MHAD: \textit{wave, hand catch, right arm high throw, draw x, draw circle, two hand front clap, jogging, tennis swing, tennis serve}, and \textit{pickup and throw}. There are 7 shared actions between MSRAction3DExt and MAD: \textit{wave, forward punch, throw, forward kick, side kick, jogging}, and \textit{tennis swing forehand}. The local HON4D~\cite{Oreifej2013} feature is used for the cross-dataset action recognition tasks. We extract local HON4D descriptors around 15 skeleton joints by following the process similar to~\cite{Oreifej2013}. The selected joints include head, neck, left knee, right knee, left elbow, right elbow, left wrist, right wrist, left shoulder, right shoulder, hip, left hip, right hip, left ankle, and right ankle. We use a patch size of $24 \times 24 \times 4$ for depth map with resolution of $320 \times 240$ and $48 \times 48 \times 4$ for depth map with resolution of $640 \times 480$
, then divide the patches into a $3 \times 3 \times 1$ grid. 
Since most of the real world applications of action recognition are required to recognize unseen data in the target domain, we further divide the target domain into training and test sets using cross-subject protocol, where half of the subjects are used as training and the rest subjects are used as test when a dataset is evaluated as target domain. Note that the target training set is also unlabeled. For the free parameters, we set $k=100$ and $\beta=0.01$. To avoid overfitting to the target training set, we set $T=1$ in action recognition tasks. LibLINEAR~\cite{Fan2008} is used for action recognition by following the original paper~\cite{Oreifej2013}. 

\vspace{-1.5em}
\subsubsection{Results and Discussion}
\vspace{-0.5em}
The results on three types of real world cross domain (object, digit, and action) datasets are shown in Table~\ref{tab:object},~\ref{tab:digit},and~\ref{tab:action}. The JGSA primal represents the results of JGSA method on original data space, while the JGSA linear and JGSA RBF represent the results with linear kernel and RBF kernel respectively. We follow JDA to report the results on digit datasets in the original feature space. For the action recognition task, it is hard to do eigen decomposition in the original space due to the high dimensionality, hence, the results are obtained using linear kernel. It can be observed that JGSA outperforms the state-of-the-art domain adaptation methods on most of the datasets.  
As mentioned, the general drawback of subspace centric approach is that the distribution shifts between domains are not explicitly reduced. 
The data centric methods reduce the distribution shift explicitly. However, a unified transformation may not exist to both reduce distribution shift and preserve the properties of original data. Hence, JGSA outperforms both subspace centric and data centric methods on most of the datasets. We also compare the primal and kernelized versions of the algorithm on the object recognition task (Table~\ref{tab:object}). The results show that the primal and kernelized versions can obtain similar results on average.
To evaluate the effectiveness of pseudo labelling, we compare our method with a semi-supervised method KEMA~\cite{Tuia2016}. We use the same $Decaf_7$ feature on 8 Office-Caltech dataset pairs as did in KEMA. Our method obtains 90.18\% (linear) and 89.91\% (RBF), both of which are higher than 89.1\% reported in KEMA. 

We also evaluated the runtime complexity on the cross-domain object datasets (SURF with linear kernel). The average runtime is 28.97s, which is about three times as long as the best baseline method (JDA). This is because JGSA learns two mappings simultaneously, the size of matrix for eigen decomposition is doubled compared to JDA.
\vspace{-1em}
\subsubsection{Parameter Sensitivity}
\vspace{-0.5em}
We analyse the parameter sensitivity of JGSA on different types of datasets to validate that a wide range of parameter values can be chosen to obtain satisfactory performance. The results on different types of datasets have validated that the fixing $\lambda=1$ and $\mu=1$ is sufficient for all the three tasks. Hence, we only evaluate other three parameters ($k$, $\beta$, and $T$). We conduct experiments on the USPS$\rightarrow$MNIST, W$\rightarrow$A (SURF descriptor with linear kernel), and MSR$\rightarrow$MAD datasets for illustration, which are shown in Figure~\ref{fig:params}. The solid line is the accuracy on JGSA using different parameters, and the dashed line indicates the results obtained by the best baseline method on each dataset. Similar trends are observed on other datasets. 

$\beta$ is the trade-off parameter of within and between class variance of source domain. If $\beta$ is too small, the class information of source domain is not considered. If $\beta$ is too big, the classifier would be overfit to the source domain. However, it can be seen from Figure~\ref{fig:beta}, a large range of $\beta$ ($\beta\in [2^{-15},0.5]$) can be selected to obtain better results than the best baseline method.

Figure~\ref{fig:k} illustrates the relationship between various k and the accuracy.
We can choose $k\in [20,180]$ to obtain better results than the best baseline method.

For the number of iteration T, the results on object and digit recognition tasks can be converged to the optimum value after several iteration. However, for the action recognition, the accuracy has no obvious change (Figure~\ref{fig:T}). 
This may be because we use a different protocol for action recognition as mentioned in Section~\ref{sec:setup}. After iterative labelling (which is done on the target training set), the mappings may be sufficiently good for fitting the target training set, but it is not necessarily the case for the test set.


\vspace{-0.5em}
\section{Conclusion}
\vspace{-0.5em}
In this paper, we propose a novel framework for unsupervised domain adaptation, referred to as Joint Geometrical and Statistical Alignment (JGSA). JGSA reduces the domain shifts by taking both geometrical and statistical properties of source and target domain data into consideration and exploiting both shared and domain specific features. Comprehensive experiments on synthetic data and three different types of real world visual recognition tasks validate the effectiveness of JGSA compared to several state-of-the-art domain adaptation methods.


%
\begin{small}
\bibliographystyle{IEEEtran} 
\bibliography{CrossDataset}

\begin{thebibliography}{10}
\providecommand{\url}[1]{#1}
\csname url@samestyle\endcsname
\providecommand{\newblock}{\relax}
\providecommand{\bibinfo}[2]{#2}
\providecommand{\BIBentrySTDinterwordspacing}{\spaceskip=0pt\relax}
\providecommand{\BIBentryALTinterwordstretchfactor}{4}
\providecommand{\BIBentryALTinterwordspacing}{\spaceskip=\fontdimen2\font plus
\BIBentryALTinterwordstretchfactor\fontdimen3\font minus
  \fontdimen4\font\relax}
\providecommand{\BIBforeignlanguage}[2]{{%
\expandafter\ifx\csname l@#1\endcsname\relax
\typeout{** WARNING: IEEEtran.bst: No hyphenation pattern has been}%
\typeout{** loaded for the language `#1'. Using the pattern for}%
\typeout{** the default language instead.}%
\else
\language=\csname l@#1\endcsname
\fi
#2}}
\providecommand{\BIBdecl}{\relax}
\BIBdecl

\bibitem{Pan2010}
S.~J. Pan and Q.~Yang, ``A survey on transfer learning,'' \emph{IEEE
  Transactions on Knowledge and Data Engineering}, vol.~22, no.~10, pp.
  1345--1359, 2010.

\bibitem{Shao2015}
L.~Shao, F.~Zhu, and X.~Li, ``Transfer learning for visual categorization: A
  survey,'' \emph{IEEE transactions on neural networks and learning systems},
  vol.~26, no.~5, pp. 1019--1034, 2015.

\bibitem{Ben-David2010}
S.~Ben-David, J.~Blitzer, K.~Crammer, A.~Kulesza, F.~Pereira, and J.~W.
  Vaughan, ``A theory of learning from different domains,'' \emph{Machine
  learning}, vol.~79, no. 1-2, pp. 151--175, 2010.

\bibitem{Margolis2011}
A.~Margolis, ``A literature review of domain adaptation with unlabeled data,''
  Tech. Rep., 2011.

\bibitem{Yang2015}
Y.~Yang and T.~Hospedales, ``Zero-shot domain adaptation via kernel regression
  on the grassmannian,'' in \emph{Proc. the 1st International Workshop on
  DIFFerential Geometry in Computer Vision for Analysis of Shapes, Images and
  Trajectories}.\hskip 1em plus 0.5em minus 0.4em\relax BMVA Press, September
  2015, pp. 1.1--1.12.

\bibitem{Pan2011}
S.~J. Pan, I.~W. Tsang, J.~T. Kwok, and Q.~Yang, ``Domain adaptation via
  transfer component analysis,'' \emph{IEEE Transactions on Neural Networks},
  vol.~22, no.~2, pp. 199--210, 2011.

\bibitem{Long2013}
M.~Long, J.~Wang, G.~Ding, J.~Sun, and P.~Yu, ``Transfer feature learning with
  joint distribution adaptation,'' in \emph{Proc. IEEE International Conference
  on Computer Vision}.\hskip 1em plus 0.5em minus 0.4em\relax IEEE, 2013, pp.
  2200--2207.

\bibitem{Long2014}
M.~Long, J.~Wang, G.~Ding, J.~Sun, and P.~S. Yu, ``Transfer joint matching for
  unsupervised domain adaptation,'' in \emph{Proc. IEEE Conference on Computer
  Vision and Pattern Recognition}.\hskip 1em plus 0.5em minus 0.4em\relax IEEE,
  2014, pp. 1410--1417.

\bibitem{Ghifary2016}
M.~Ghifary, D.~Balduzzi, W.~B. Kleijn, and M.~Zhang, ``Scatter component
  analysis: A unified framework for domain adaptation and domain
  generalization,'' \emph{IEEE Transactions on Pattern Analysis and Machine
  Intelligence}, vol.~PP, no.~99, pp. 1--1, 2016.

\bibitem{Gong2012}
B.~Gong, Y.~Shi, F.~Sha, and K.~Grauman, ``Geodesic flow kernel for
  unsupervised domain adaptation,'' in \emph{Proc. IEEE Conference on Computer
  Vision and Pattern Recognition}, 2012, pp. 2066--2073.

\bibitem{Fernando2013}
B.~Fernando, A.~Habrard, M.~Sebban, and T.~Tuytelaars, ``Unsupervised visual
  domain adaptation using subspace alignment,'' in \emph{Proc. IEEE
  International Conference on Computer Vision}, 2013, pp. 2960--2967.

\bibitem{Fernando2015}
B.~Fernando, T.~Tommasi, and T.~Tuytelaars, ``Joint cross-domain classification
  and subspace learning for unsupervised adaptation,'' \emph{Pattern
  Recognition Letters}, vol.~65, pp. 60--66, 2015.

\bibitem{Gretton2012}
A.~Gretton, K.~M. Borgwardt, M.~J. Rasch, B.~Sch{\"o}lkopf, and A.~Smola, ``A
  kernel two-sample test,'' \emph{Journal of Machine Learning Research},
  vol.~13, pp. 723--773, 2012.

\bibitem{Sun2015}
B.~Sun and K.~Saenko, ``Subspace distribution alignment for unsupervised domain
  adaptation,'' in \emph{Proc. British Machine Vision Conference}, 2015.

\bibitem{Courty2016}
N.~Courty, R.~Flamary, D.~Tuia, and A.~Rakotomamonjy, ``Optimal transport for
  domain adaptation,'' \emph{IEEE Transactions on Pattern Analysis and Machine
  Intelligence}, vol.~PP, no.~99, pp. 1--1, 2016.

\bibitem{Tuia2016}
D.~Tuia and G.~Camps-Valls, ``Kernel manifold alignment for domain
  adaptation,'' \emph{PloS one}, vol.~11, no.~2, p. e0148655, 2016.

\bibitem{Saenko2010}
K.~Saenko, B.~Kulis, M.~Fritz, and T.~Darrell, ``Adapting visual category
  models to new domains,'' in \emph{European conference on computer
  vision}.\hskip 1em plus 0.5em minus 0.4em\relax Springer, 2010, pp. 213--226.

\bibitem{Griffin2007}
G.~Griffin, A.~Holub, and P.~Perona, ``Caltech-256 object category dataset,''
  Tech. Rep., 2007.

\bibitem{LeCun1998}
Y.~Lecun, L.~Bottou, Y.~Bengio, and P.~Haffner, ``Gradient-based learning
  applied to document recognition,'' \emph{Proceedings of the IEEE}, vol.~86,
  no.~11, pp. 2278--2324, Nov 1998.

\bibitem{Hull1994}
J.~J. Hull, ``A database for handwritten text recognition research,''
  \emph{IEEE Transactions on pattern analysis and machine intelligence},
  vol.~16, no.~5, pp. 550--554, 1994.

\bibitem{Li2010}
W.~Li, Z.~Zhang, and Z.~Liu, ``Action recognition based on a bag of 3d
  points,'' in \emph{Proc. IEEE Computer Society Conference on Computer Vision
  and Pattern Recognition Workshops}.\hskip 1em plus 0.5em minus 0.4em\relax
  IEEE, 2010, pp. 9--14.

\bibitem{Wang2016}
P.~Wang, W.~Li, Z.~Gao, J.~Zhang, C.~Tang, and P.~O. Ogunbona, ``Action
  recognition from depth maps using deep convolutional neural networks,''
  \emph{IEEE Transactions on Human-Machine Systems}, vol.~46, no.~4, pp.
  498--509, Aug 2016.

\bibitem{Chen2015b}
C.~Chen, R.~Jafari, and N.~Kehtarnavaz, ``{UTD-MHAD}: A multimodal dataset for
  human action recognition utilizing a depth camera and a wearable inertial
  sensor,'' in \emph{Proc. IEEE International Conference on Image Processing},
  2015.

\bibitem{Bloom2012}
V.~Bloom, D.~Makris, and V.~Argyriou, ``{G3D}: A gaming action dataset and real
  time action recognition evaluation framework,'' in \emph{Proc. IEEE Computer
  Society Conference on Computer Vision and Pattern Recognition Workshops},
  June 2012, pp. 7--12.

\bibitem{Huang2014}
D.~Huang, S.~Yao, Y.~Wang, and F.~De~La~Torre, ``Sequential max-margin event
  detectors,'' in \emph{European conference on computer vision}.\hskip 1em plus
  0.5em minus 0.4em\relax Springer, 2014, pp. 410--424.

\bibitem{Oreifej2013}
O.~Oreifej and Z.~Liu, ``{HON4D}: Histogram of oriented {4D} normals for
  activity recognition from depth sequences,'' in \emph{Proc. IEEE Conference
  on Computer Vision and Pattern Recognition}.\hskip 1em plus 0.5em minus
  0.4em\relax IEEE, 2013, pp. 716--723.

\bibitem{Fan2008}
R.-E. Fan, K.-W. Chang, C.-J. Hsieh, X.-R. Wang, and C.-J. Lin, ``{LIBLINEAR}:
  A library for large linear classification,'' \emph{Journal of machine
  learning research}, vol.~9, no. Aug, pp. 1871--1874, 2008.

\end{thebibliography}
\end{small}
\end{document}